# A Hierarchical Fuzzy System for an Advanced Driving Assistance System

Mejdi Ben Dkhil, Ali Wali, and Adel M. Alimi
*REsearch Groups in Intelligent Machines University
of Sfax, National School of Engineers (ENIS) BP
1173, 3038 Sfax, Tunisia*

{mejdi.bendkhil, ali.wali, adel.alimi}@ieee.org

*Abstract*— **In this study, we present a hierarchical fuzzy system by evaluating the risk state for a Driver Assistance System in order to contribute in reducing the road accident's number. A key component of this system is its ability to continually detect and test the inside and outside risks in real time: The outside car risks by detecting various road moving objects; this developed system is stand on computer vision techniques. The inside risks by presenting an automatic system for drowsy driving identification or detection by analyzing EEG signals of the driver; this developed system is based on computer vision techniques and biometrics factors (electroencephalogram EEG). This proposed system is then composed of three main modules. The first module is responsible for identifying the driver drowsiness state through his eye movements (physical drowsiness). The second one is responsible for detecting and analysing his physiological signals to also identify his drowsiness state (moral drowsiness). The third module is responsible to evaluate the road driving risks by detecting of the road different moving objects in a real time. The final decision will be obtained by merging of the three detection systems through the use of fuzzy decision rules. Finally, the proposed approach has been improved on ten samples from a proposed dataset.**

## I. INTRODUCTION

Traffic accidents are the major problem in which millions of people die. In literature, we have improved that there are two types of risks: Driver drowsiness is the first factor for the big developing number of vehicles accidents [1] has improved that driving drowsiness increase from four to six times the risk of having an accident. In fact, each year, up to 20% of car accidents are the result of the drowsiness of a driver [2]. Drowsiness refers to the passage from being awake to feeling asleep in which an individual's capability to master and evaluate become highly decreased. This state boosts in reaction time along with a decrease in the vivacity of the driver which result in a damage of driving abilities. Capabilities to boost traffic security draw the attention of the researchers to concentrate on detecting this alarming situation in a systematic way. Drowsiness can be evaluated by studying the behavior of the driver, employing data from sensors placed in the vehicle like the road position, the movements of the steering wheel, and stress on the driving pedals or the changes in the vehicle speed's. The ultimate drawback of this method is that driving behavior can vary among drivers. Therefore, it is crucial to create a "correct driving model" applied to evaluate differences in the behaviors of drivers and hence, it must be learnt by drivers. Based on the system type two known as the "driver-oriented" [3], drowsiness can be recognized through the use of physiological data derived from sensors placed in or around the driver (electrodes, camera, etc.) as cerebral activity, eye activity, facial expressions, yawns, or gaze direction. These driver-oriented techniques are authentic as physiological drowsiness signs, prominent, as well as common among drivers [4]. Yet, placing sensors on the driver has a drawback as it can bother him/her. Furthermore, quantification can be challenging as the individual is in a constant movement. According to [5] ,physiologically speaking, drowsiness can be defined as employing both brain activity, which indicates the brain's capability to analyze the data, and visual blinks, that reflects an indication of perceptive capability. Thusly, the data applied by expert doctors to determine drowsy person's level [6] Other studies pay considerable interest in the study of visual signs, like blinking or visual gaze, to evaluate the drowsiness level of the driver. In the study [7] points out that excellent clues of drowsiness are derived from blinking investigations [8]. Based on this work, a system that detects the drowsiness level employs mental as well as visual signs[9]. Two detection techniques are suggested. The first system makes use of the differences of activity in various frequency fields calculated by one EEG channel whereas the second system offers a few blinking characteristics taken from video. In [10] authors' demonstrates that the features applied are restricted to features which can be selected with a high frame video analysis with the same accuracy. Henceforward, performances are predicted to be alike when a video with high frame is applied aimed at controlling eye blinking. The second factor is outside risk, the detection or prevention of moving object while driving represents a challenging choice for accident preventing methods. The fact of giving correct and accurate vehicle detection for visual sensors remains a challenging step due to the diversity of shapes, dimensions, and hues portraying the on road vehicles. Moving object detection by definition is the fact of identifying the physical movement of an object in a specific area. Hence, for this serious risk, we suggest to develop a system which, both, supervises the road moving object and estimates the driver drowsiness level, in a given time. This system is highly recommended for drivers, particularly when they are tired, to minimize car accidents. The rest of the paper is based on three basic sections as follows: The second section presents the Material and databases. The third section presents the experimental results and discussions of this work. Finally, it presents the conclusion and some suggestions for future researches.

## II. METHODS

In this section, two smart cameras are used, the first is used for eye movement detection and the second is used for the street moving object detection, and an EPOC headset for EEG measurement.

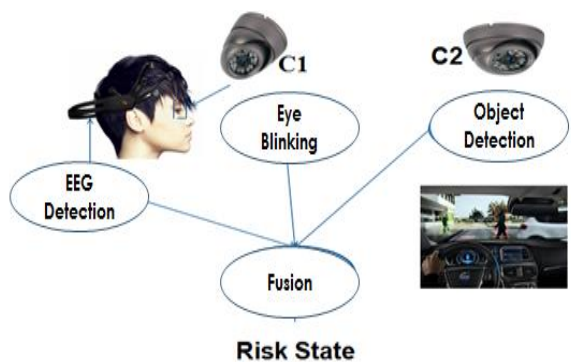

Fig.1. Architecture of the proposed control

### A. Material

Because there is no datasets that contains three different measurements, we propose to develop one. The database implemented in this study was supplied by the Higher Institute of Transport and Logistics (ISTLS) from Sousse, Tunisia, with collaboration with the REsearch Groups in Intelligent Machines (REGIM) from Sfax, Tunisia. The Regim_EBR (The Regim Eye Brain and Road) is the proposed database used in this work. In the Regim_EBR [11], 10 different test candidates are subject in this database. During one minute, for each person, while he is watching in a KITTI sequence of road moving object for one minute, we use a cam to record his eye movements, and we use EPOC headset to record his EEG signals. The database is composed of 20 recordings from 10 individuals. Everyone was recorded while driving for 1 min, in the first step while completely relaxed and in the second step endure sleeping need (the individual had just 2 h of sleep in the previous night) in normal situations.

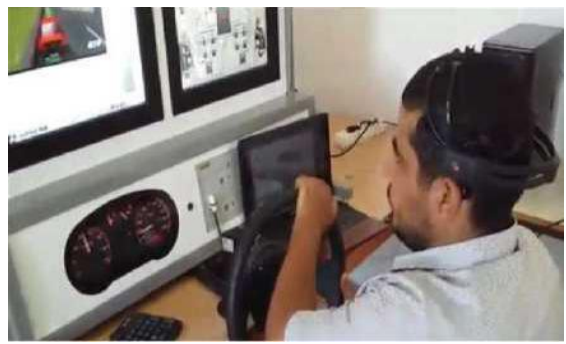

Fig.2. A subject example from The Regim_EBR database

### A. Drowsiness detection based on eye blinking

In our proposed system, we have fixed a smart camera to the dashboard of car. It takes images of various states of driver's drowsiness detection.

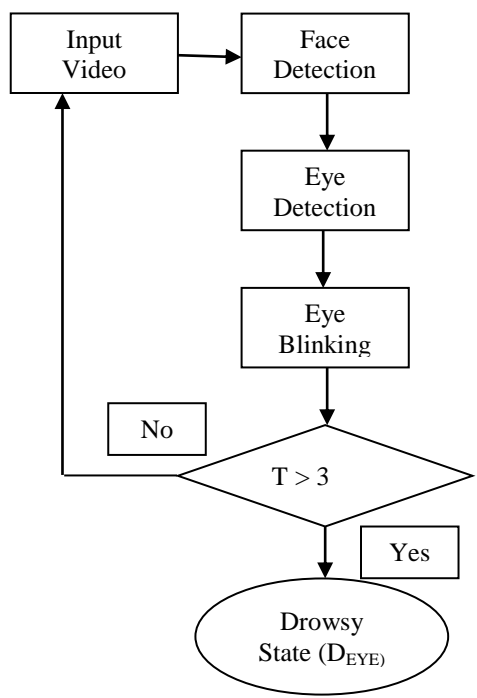

Fig.3. Proposed system for detecting driver drowsiness based on eye blinking analyzing

In our system, face and eyes can be detected by the method of Viola&Jones [12]. This method allows the detection of objects for which learning was achieved. It was designed specifically for the purpose of face detection, but may also be used for other types of objects. As a supervised learning method, the method of Viola&Jones requires hundreds to thousands of examples of the detected object to train a classifier. The classifier is then used in an exhaustive search of the object for all possible positions and sizes the image to be processed. The detection of eye blinking in real time is very important to estimate driver drowsiness state.

Based on the literature, the PERCLOS (Percentage of eye Closure) [13] value has been used as drowsiness metric which shows the percentage of closure in specific time (e.g in a minute, eyes are 80% closed). Using these eyes closer and blinking ration, one can detect drowsiness of driver. Then, we move to the following frame until obtaining closed eyes. We calculate the duration of eye closure; if it exceeds a predefined Time T (T = 3 seconds), we may note that the driver enters in a drowsiness state.

In this work, we note:

- IF (T= 0 sec), we note T is SMALL.
- IF (0<T<3 sec), we note T is MEDIUM.
- IF (T>3 sec), we note T is BIG.

We attempt to validate our suggested system which controls risk level by defining fuzzy rules:

1. IF (T is SMALL) THEN ($D_{EYE}$ is SMALL).
2. IF (T is MEDIUM) THEN ($D_{EYE}$ is MEDIUM).
3. IF (T is BIG) THEN ($D_{EYE}$ is BIG).

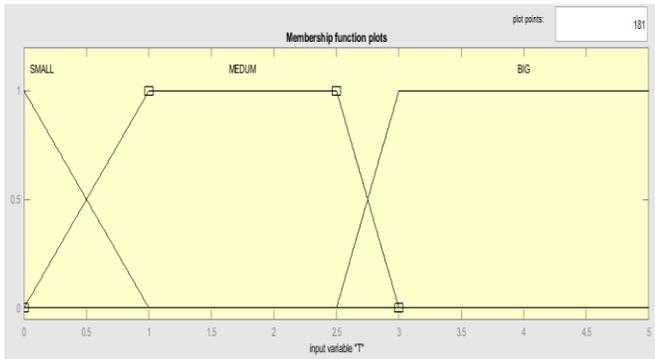

Fig.4. Characterization of T in fuzzy logic.

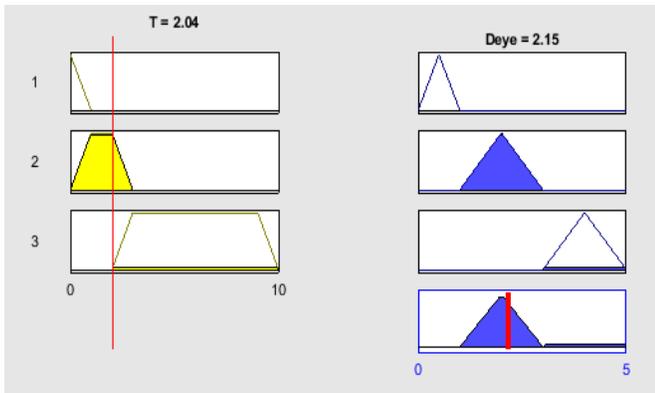

Fig.5. Fuzzy distributions of T and $D_{EYE}$.

## B. Drowsiness detection by evaluating the EEG signals

The EEG recorded in the absence of external stimulus is called Spontaneous EEG. The EEG generated further to an answer to an internal or external stimulus is called VRP (Vent Related Potential). The EEG can be measured by using electrodes placed at the level of the skin. The location of every electrode is standardized by an international list called system 10-20 [14]. This system is symmetric; its purpose is to place the neighboring electrodes in an equidistant way frontally behind and from right to left. They are placed in 10 % or 20 % of the total distance and among which each possesses the own name with P: parietal bone, O: occipital, F: frontal bone, Fp: frontal-parietal bone, T: temporal and C: power plant [15].

In [16], authors describe the signal EEG as being the contribution of the intellectual activities Rhythmic which are divided into frequency bands. Let us add to it that these rhythmic activities depend on the age and on the behavior in state of vigilance. Let us give now an overview onto the various frequency bands characterizing the signal EEG:

- The Delta activity (0.5-4 hz): this activity is recorded during the sound sleep. At this stage the wave delta has strong amplitude (of the order of 75-200mvs).
- The Theta activity (4-8 hz): it rarely expresses himself at the adult's however she is dominant at the child's and at the young people during the half-sleep.
- The Alpha activity (8-13 hz): they are dominant during the awakening in the later regions of the head. They are generally better observed when we close eyes.
- The Beta activity (12-30 hz): it characterizes mainly the awakening, the increase of the vigilance and the attention.
- The Gamma activity (30-50 hz): it appears during certain cognitive spots or driving functions. It is difficult to record in standard clinical conditions.

Based on the literature, effective systems have been explored using EEG in order to control the drowsiness of a driver. Drowsiness is characterized by the differences in spectral power in determent frequency waves of the activity of the brain: The length of these eruptions is applied to determinate drowsiness. These changes are investigated by analyzing a spectrum analysis of the EEG signal. By looking at the meaning of the various bands establishing the signal EEG, we can conclude that the hypovigilance is mainly characterized by the activities theta (which correspond to the half-sleep), alpha (which corresponds to the closure of eyes), and finally, Beta (which corresponds to the awakening and to the increase of the vigilance).

- Feature extraction

  - Arousal: It is distinguished by a high beta power and coherence in the parietal lobe as well as low alpha activity. Beta waves are connected to a state of an alert or excited mind, while alpha waves are more dominant in a state of relaxation. Thus the beta/alpha ratio is a reasonable indicator of the excitation state of a person.

$$\text{Arousal} = \frac{\alpha(AF3 + AF4 + F3 + F4)}{\beta(AF3 + AF4 + F3 + F4)} \quad (1)$$

- Valence: The prefrontal lobe (F3 and F4) plays a crucial role in the regulation of drowsiness and conscious experience.

$$\text{Valence} = \frac{\alpha F4}{\beta F4} - \frac{\alpha F3}{\beta F3} \quad (2)$$

- Dominance: It is characterized by an increase in the ratio (beta/alpha) activity in the frontal lobe and an increase in the beta activity in the parietal lobe.

$$\text{Dominance} = \frac{\beta FC6}{\alpha FC6} + \frac{\beta F8}{\alpha F8} + \frac{\beta P8}{\alpha P8} \quad (3)$$

To extract the arousal, valence and dominance drowsiness features, we present the algoithm1.

**Algorithm1: Drowsiness detection**

**Inputs:** EEG signals.
**Output:** Drowsy state.
**Begin**
  Loading a CSV file.
  **For** each data of the signal (treatment of 20 s of the signal) **do**
  - Applying the FFT filter on the signal of electrodes
  - Construction of the two pass-band filter Alpha and Beta.
  - Applying the pass-band filter on the different electrodes.
  - Computing the values of Arousal, valence and dominance

  **End For**
  Clustering by Fuzzy Cmeans.
**End.**

➢ Fuzzy Logic Classification

In this part, we tend to classify the EEG signals. We apply the fuzzy logic techniques, among which the "Mamdani" one is used. We have three inputs: arousal, valence, dominance and the state of drowsiness is the output variable.

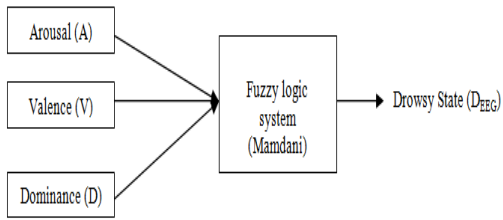

Fig.6. Fuzzy logic system,

We define for each input variable three membership functions: SMALL, MEDIUM and BIG. We propose in what follows the fuzzy rules:

Table.1. Inference matrix of the EEG detection system using a logical "AND"

| VALENCE | AROUSAL | Dominance | $D_{EEG}$ |
|---------|---------|-----------|-----------|
| BIG | * | * | BIG |
| MEDUIM | SMALL | SMALL | MEDIUM |
| MEDUIM | SMALL | MEDIUM | MEDIUM |
| MEDUIM | MEDIUM | SMALL | MEDIUM |
| MEDUIM | MEDIUM | MEDIUM | MEDIUM |
| SMALL | SMALL | SMALL | SMALL |
| SMALL | SMALL | MEDIUM | SMALL |
| SMALL | MEDIUM | SMALL | SMALL |
| SMALL | MEDIUM | MEDIUM | MEDIUM |

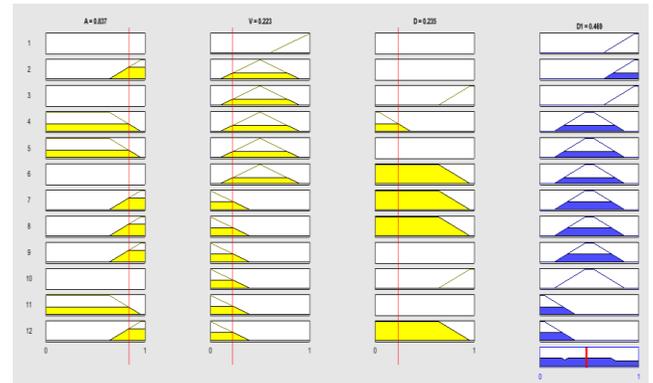

Fig.7. Fuzzy distributions of Arousal, Valence, Dominance and $D_{EEG}$.

*B. Road moving object detection system*

In our proposed system, we have fixed a smart camera to the dashboard of car. It takes images of different road moving objects (pedestrians, cars, cyclists, pets, etc.).

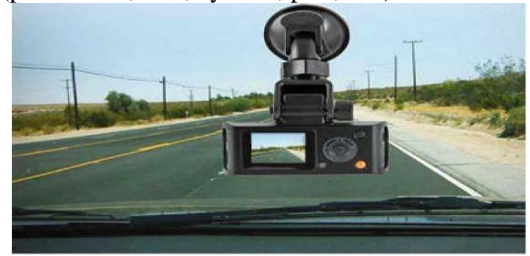

Fig.8. System overview

According to [17] the flowchart of the architecture system is presented in Fig.9.

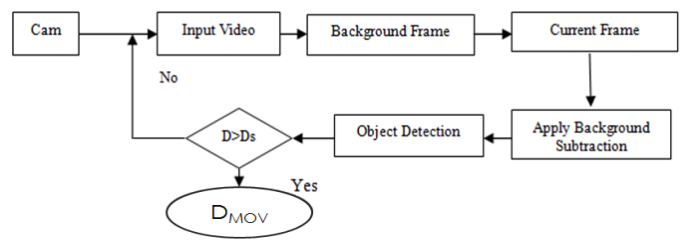

Fig.9. Architecture of the road detection moving object

We tend to prove our proposed system that controls risk level by calculating the distance to stop Ds [18]. We note:

1. Ds: Distance to stop (meters).
2. Dr: Reaction Distance (meters).
3. Db: Braking Distance (meters).
4. Tr: Reaction Time (seconds).
5. S: Speed (km/h).
6. D: the distance between our car and the detected object, it is given by the computer's calculator.
7. Tr = 1 second for a vigilant person.
8. θ = 1 in better weather and θ = 1.5 in runny weather.

We evaluate this distance Dr by formulas:

$$Dr = Tr * \frac{S*1000}{3600} \qquad (4)$$

$$Db = \frac{S*3}{10} * θ \qquad (5)$$

$$Ds = [(Tr * \frac{S*1000}{3600}) + (\frac{S*3}{10} * θ)] \qquad (6)$$

In this work, we note:

➢ If (D> Ds), we note D is SMALL.
➢ If (D≤ Ds) and (D> Dr)), we note D is MEDIUM.
➢ If (D≤ Dr), we note D is BIG.

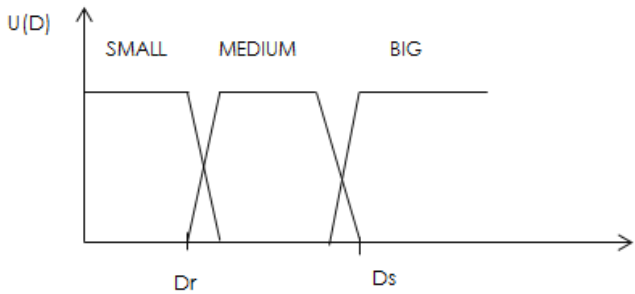

Fig.10. Characterization of D in fuzzy logic.

To control the risk state, we go to define those three fuzzy rules:

1. R1: IF (D is SMALL) THEN $D_{MOV}$ is SMALL.
2. R2: IF (D is MEDIUM) THEN $D_{MOV}$ is MEDIUM.
3. R3: IF (D is BIG) THEN $D_{MOV}$ is BIG.

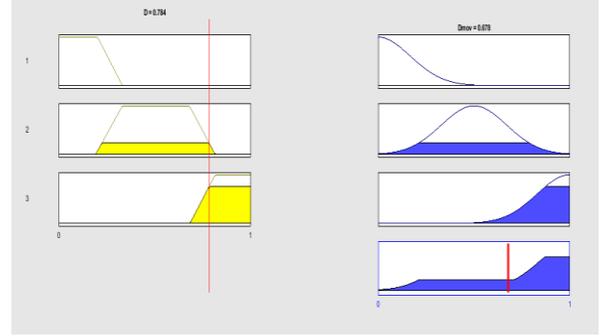

Fig.11. Fuzzy distributions of D and $D_{mov}$.

*C. Background subtraction*

With the suggested system, we can effectively change the threshold value according to the lighting changes of the two images obtained by using a Gaussian Smooth operator in order to reduce image noise and details. This method can effectively reduce the impact of light changes. Here we regard first frame as the background frame directly and then that frame is subtracted from current frame to detect moving object.

$$G(x) = \frac{1}{\sqrt{2\pi\sigma^2}} e^{-\frac{x^2}{2\sigma^2}} \qquad (7)$$

The Viola-Jones algorithm applies the Haar-like features[19]. The comprehensive search for an item is inside an image which can be measured in computing time. Every classifier decides the vicinity or nonappearance of the item in the image. The least difficult and quickest classifiers are put in the first place, which quickly disposes of numerous negative (Fig.12).

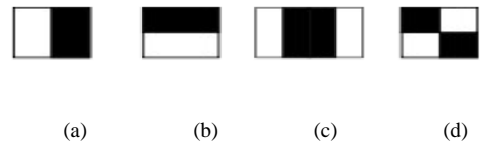

(a)　　　(b)　　　(c)　　　(d)

Fig.12. Examples of neighborhoods used

Broadly speaking, the technique of Viola&Jones presents effective results in the Face Detection or different articles, with few false positives for a low figuring time, permitting the operation here progressively. The recognition of different road moving objects is highly crucial to reduce the occurrence of an accident.

## D. Risk Architecture system

It can be noted in this work, that risk appears on three ways: first in physical activity by analyzing eye blinking, second in brain activity by calculating the alpha band of EEG signals and third, an outside car's risk by analyzing the road moving object. For both the first and the second way of risks: drowsiness is marked with cerebral and visual signals. For the third way of risk: road moving objects are detected to evaluate risk state in the instant t.

Three systems to evaluate risk from varied source have been mentioned in previous section. A decision system combining the taken decisions by the three detectors is at present invented aimed at facilitating the notes: $D_{EEG}$, $D_{EYE}$ and $D_{MOV}$ are the inputs of the fuzzy system in Fig.13.

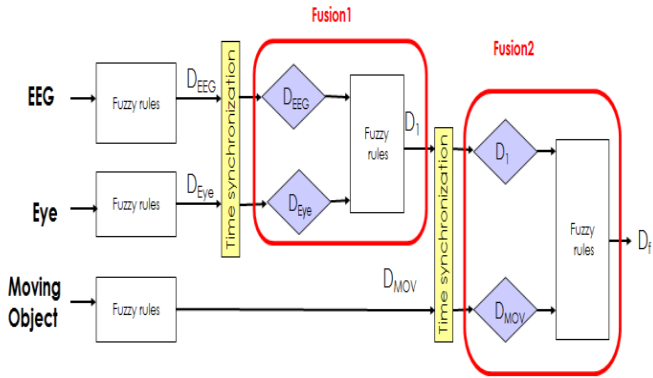

Fig.13. Hierarchical fuzzy system for the risk detection.

During the first part of the algorithm, eye blinking detection is been calculated. These detections are measured every 20s. A temporal synchronization step is measured to assure that the required information to measure the fusion step is possible. When the three detections $D_{EEG}$, $D_{EYE}$ and $D_{MOV}$ are made, they are fusion applying decision fuzzy rules.

In this work, we recommend to decompose risk state into three states:
1. Internal risk: by merging the physical activities (eye blinking analysis) and moral activities (EEG analysis) to evaluate the risk state.
2. External risk: by evaluating the road moving object detection system to evaluate the risk state.
3. Fusion of the two sub-systems to evaluate the final risk state.

➢ Drowsiness fuzzy system

Two systems used to evaluate drowsiness from two various parts are shown in the preceding sections. A decision system fusion the decisions made by the two detectors is now presented. The fusion system is positively affected by the OSS scale that is adopted by expert doctors to estimate the level of drowsiness [20]. This scale is shown in table 2.

Table.2. Data evaluated every 20s according to the OSS scale

| Objective sleepiness score | EEG activity on the Alpha and Beta bands | Eye blinking |
|---|---|---|
| 0 | Negligible | Normal |
| 1 | Alpha less than 5 seconds | Normal |
| 2 | Alpha less than 5 seconds | Slow |
| 3 | Alpha less than 10 seconds | Slow |
| 4 | Alpha less than 10 seconds | Slow |

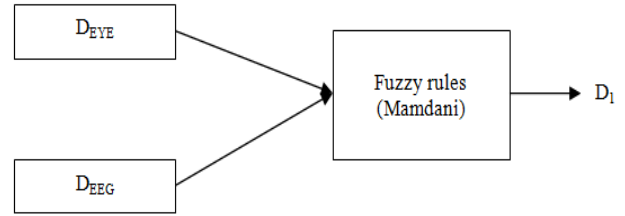

Fig.14. Drowsiness fuzzy system architecture

The proposed algorithm for evaluating drowsiness state adopting both EEG and EYE blinking information is defined by fuzzy rules given in Table.3:

Table.3. Inference matrix of the EEG/EYE fusion detection system using a logical "AND"

| Deeg \ Deye | SMALL | MEDIUM | BIG |
|---|---|---|---|
| SMALL | SMALL | MEDIUM | BIG |
| MEDIUM | MEDIUM | MEDIUM | BIG |
| BIG | BIG | BIG | BIG |

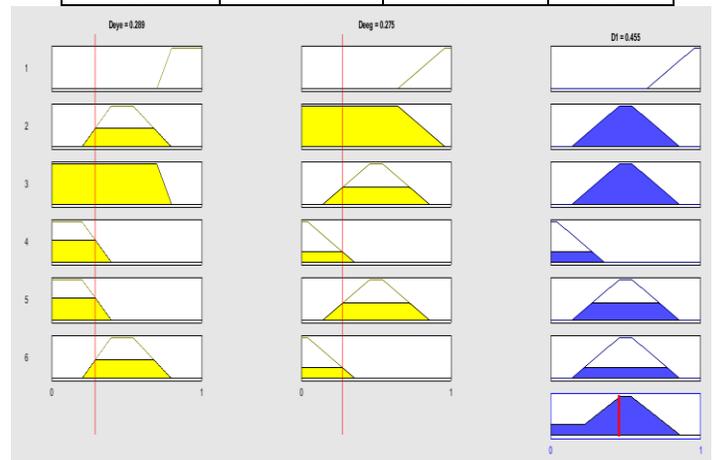

Fig.15. Fuzzy distributions of $D_{EYE}$, $D_{EEG}$ and $D_1$.

➢ Hierarchical fuzzy system

The algorithm to detect final risk state using three different sources of information (EEG signals, EYE blinking information and road moving objects) is defined by the following proposed fuzzy rules:

Table.4. Inference matrix of the hierarchical fuzzy system using a logical "AND"

| $D_{MOV}$ \ $D_1$ | SMALL | MEDIUM | BIG |
|---|---|---|---|
| SMALL | SMALL | MEDIUM | BIG |
| MEDIUM | MEDIUM | MEDIUM | BIG |
| BIG | BIG | BIG | BIG |

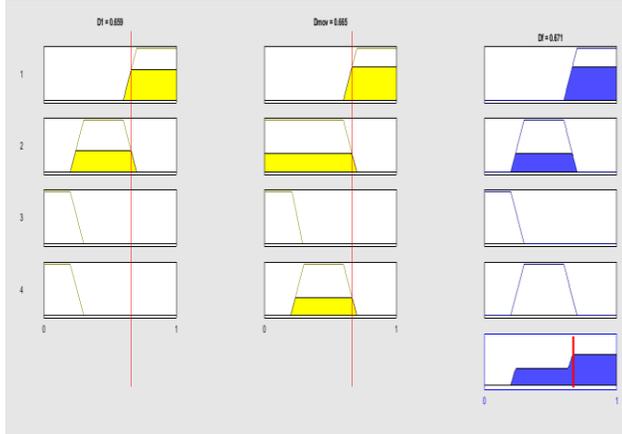

Fig.16. Fuzzy distributions of $D_1$, Dmov and $D_f$.

## III. RESULTS AND DISCUSSION

In this part, we present different experimental results elaborated throughout this work. The first system describes drowsiness obtained from the detection of eye blinking. The second system stands for detecting drowsiness from evaluating EEG signals. The third system refers to the detection of road moving objects. Finally, a hierarchical fuzzy system combines the three decisions that will be proved by the proposed database.

The results are obtained applying confusion matrices. The ratios are calculated in every single column. The last row show the epoch's number in the validation set mentioned in different classes.

$D_{EEG}$: The drowsiness detection system depends on the investigation of the EEG signals.

1. $D_{EYE}$: The drowsiness detecting system depending on eye blinking measurement.

2. $D_{EEG}$: The drowsiness detecting system depending on the analysis of EEG signals.

3. $D_1$: The drowsiness detection system based on fusion EYE information and EEG signals information.

4. $D_{MOV}$: The road moving object detection system.

5. Fusion System ($D_f$): The results obtained by the merge of the three sub-systems: $D_{EEG}$, $D_{EYE}$ and $D_{MOV}$ in order to calculate the risk level in a given instant.

We are going to represent different experimental results as well as three approaches developed in this section.

The results are represented applying confusion matrices. The columns are based on the expert's classification and the rows to the classification are completed by the technique. The ratios can be measured on each column. The epochs' number in the validation set maintained in each class is displayed in the last row.

The objective tends to prove a favorable compromise between lots of correct detections of "drowsy" epochs and a low number of false alarms. In fact, insufficient correct detections of "drowsy" epochs represent a risk for the driver. However, a great number of false alarms may restrict the potential of the system and the driver will not observe the true alarms.

### A. Eye blinking based detection system

As a result of the analysis procedure, the detecting drowsiness system based on the EYE blinking analysis attains 79.46% correct detections of "big drowsy" cases resembling to drowsiness phase higher or equal to 2 in the OSS scale (table.5) and 15.46% false alarms (11.93 % categorized epoch 1: "drowsy: MEDIUM" by the expert and evaluated as "drowsy: MEDIUM". and 3.53% epochs classed 0: "awake: SMALL" by the expert and evaluated as "very drowsy: BIG" by the EYE blinking related to the detection system.

Table.5. Confusion matrix of the EYE blinking-based method

| | EXPERT | | |
|---|---|---|---|
| | SMALL (level 0) | MEDIUM (level 1) | BIG (level ≥ 2) |
| SMALL | 86.05% | 16.64% | 8.01% |
| MEDIUM | 11.42% | 71.53% | 12.53% |
| BIG | 3.53% | 11.93% | 79.46% |

| SAMPLES | 730 | 244 | 160 |
|---|---|---|---|

### B. EEG based detection system

As a result of the analysis procedure, the detecting drowsiness system based on the EEG analysis attains 76.98% correct detections of "big drowsy" cases resembling to drowsiness phase higher or equal to 2 in the OSS scale (table.6) and 15.44% false alarms (10.31 % categorized epoch 1: "drowsy: MEDIUM" by the expert and evaluated as "drowsy: MEDIUM". and 4.83% epochs classed 0: "awake: SMALL" by the expert and evaluated as "very drowsy: BIG" by the EEG analysis related to the detection system.

Table.6. Confusion matrix of the EEG-based method

| | Expert | | |
|---|---|---|---|
| | SMALL (level 0) | MEDIUM (level 1) | BIG (level ≥ 2) |
| SMALL | 82.15% | 15.22% | 5.19% |
| MEDIUM | 13.02% | 74.17% | 17.83% |
| BIG | 4.83% | 10.61% | 76.98% |

| SAMPLES | 867 | 202 | 65 |
|---|---|---|---|

*C. Drowsiness system detection based on EEG/EYE fusion*

As a result of the analysis procedure, the detecting drowsiness system based on the EEG analysis attains 78. 36% correct detections of "big drowsy" cases resembling to drowsiness phase higher or equal to 2 in the OSS scale (table.7) and 9.76% false alarms (6.23 % categorized epoch 1: "drowsy: MEDIUM" by the expert and evaluated as "drowsy: MEDIUM", and 3. 53% epochs classed 0: "awake: SMALL" by the expert and evaluated as "very drowsy: BIG" by the fusion of EEG/EYE system.

Table.7. Confusion matrix of the fusion of EEG/EYE-based method

| | Expert | | |
|---|---|---|---|
| | SMALL (level 0) | MEDIUM (level 1) | BIG (level ≥ 2) |
| SMALL | 86.05% | 16.64% | 8.01% |
| MEDIUM | 11.42% | 76.13% | 13.63% |
| BIG | 3.53% | 6.23% | 78.36% |

| SAMPLES | 730 | 244 | 160 |
|---|---|---|---|

*D. Moving object based detection system*

As a result of the analysis phase, the risk detection system based on the road moving object detection attains 73.23% correct detections of "big risk" states.

Table.8. Confusion matrix of the road moving object detection-based method

| | Expert | | |
|---|---|---|---|
| | SMALL | MEDIUM | BIG |
| SMALL | 87.12% | 11.28% | 9.41% |
| MEDUIM | 10.83% | 75.23% | 16.36% |
| BIG | 2.05% | 14.49% | 74.23% |

| SAMPLES | 816 | 186 | 132 |
|---|---|---|---|

*E. Hierarchical fuzzy system*

Throughout this part, the $D_{EEG}$ stands for detecting drowsiness from evaluating EEG signals, the $D_{EYE}$ obtained from the detection of eye blinking $D_{EYE}$ and the $D_{MOV}$ refers to the detection of road moving objects, are combined and the three decisions are improved on the proposed database.

As a result of the analysis phase, The drowsiness detection system made from the EEG analysis attains 79.46% correct detections of "BIG" states, which correlates with a big risk level's and 14.11% false alarms: 11.46 % as a "MEDIUM" states which refers to a medium level of risk and 2.65% as a "SMALL" states which refers to a small level of risk. So, we can consider that our system detect risk as a 90.92% (79.46 % + 11.46%).

Table.9. Confusion matrix of the hierarchical fuzzy system

| | SMALL | MEDIUM | BIG |
|---|---|---|---|
| SMALL | 87.12% | 12.41% | 7.32% |
| MEDIUM | 10.23% | 76.13% | 12.22% |
| BIG | 2.65% | 11.46% | 79.46% |

| SAMPLES | 730 | 244 | 160 |
|---|---|---|---|

The percentage of classification of risk's evaluation is given by application of the proposed fuzzy rules mentioned in Table.9: The proposed system offers a good performance with approximately 90.92% of good classification.

*F. Discussion*

In this part, we attempt to compare the effectiveness of our proposed approach with others works in literature.

By analysing Table.9, our suggested system obtains a favourable performance with approximately 90.92% of good detection ratio. In addition, this system retains all the advantages of the algorithms of detection EEG and Video presented in the previous section. These benefits are the robustness to differences between inter-individuals, the low period of learning.

The results have been also obtained on a database of ten different drivers without having to change the settings of detection.

Our system, although more efficient, does not require a long learning (only a minute of online learning is carried out on the driver for the EEG detection). In addition, it is robust to differences, inter-individual since the results have been obtained on a ten different drivers without having to change the settings of a driver on the other. The advantage of our system is its simplicity and robustness to differences inter-individual. In effect, he was able to be applied on ten different drivers without having to change the parameters of the algorithm. In addition, its performance remains adequate with 90.92% good detections ratio. Finally, the relevance of this comparison remains relative because all these techniques have been tested on different data.

The proposed system offers a 90.92% as a good detection ratio when it considers two classes of risk level's: "No risk" and "Risk", but it offers a 79.46% as a good detection ratio when it considers three risk level's: "SMALL: no risk", "MEDIUM" and "BIG", with more performance than two classes.

Table.10. Summary of the comparison of our results with those of literature on the detection of risks by fusion of information

| Authors | Classification Results | Data | Technique | Subjects | Risks Level |
|---|---|---|---|---|---|
| Our approach | 90.92 % | Video/EEG/EYE | Hierarchical fuzzy system | 10 | 2 |
| [21] | 78% | EYE/EEG | Confusion graph | 20 | 2 |
| [22] | 87% | Video/EYE | Weight Assigned | 8 | 2 |
| [23] | 81.4% | EYE/Video | Fuzzy Rules | 20 | 3 |
| [24] | 81% | EYE | SVM | 2 | 2 |
| [25] | 75% | Video | Bayesian Networks | 10 | 2 |
| [26] | 71% | Video | theory of evidence | 2 | 2 |

## IV. CONCLUSION

We have presented an approach for a hierarchical fuzzy control system in real time throughout this work. This approach is made up by two tasks. The first aims to analyze the driver's drowsiness level, while the second aims to control any external risks. Then we try to merge the two tasks to ensure a safe conduct for the car and the driver.

This work represents a first step toward a system of detection of risks level by fusion of EEG data and video. A system of merger of three algorithms for detection of risks (drowsiness and moving objects), presented is in the form of an expert system that uses different types of data to estimate in real time the risk level in order to contribute in the security of the driver and the car.

The proposed system has been tested in set examples achieving an excellent recognition rate.

Our system obtains a good performance with approximately 90.92% of good classification. The leading advantage of this system is to maintain the advantages of the algorithms of detection EEG and Video presented in the previous sections. These benefits are the robustness to differences between inter-individuals, the low period of learning as well as the assessment of the quality of the EEG signal by the detector of the alpha band. The results have also been obtained on a database of ten different drivers without having to change the settings of detection.

It is essential to mention that our system is limited by the fact that we do not have information to validate the detection EEG. It would therefore be necessary to have a system for EEG detection more end so as to improve performance. In addition, it seems born necessary to carry out the acquisition of a data base of conduct containing the EEG signals are synchronized with video in higher frequency of way to validate our system. In effect, the results have been obtained using two rapid cameras and an EEG headset as had been the case in the previous sections. The failure of our attempt to achieve such a campaign of acquisition we have shown the difficulty of such a work.

ACKNOWLEDGMENT

The research results has received funding from the Ministry of Higher Education and Scientific Research of Tunisia under the grant agreement number LR11ES48.